\documentclass[sigconf]{acmart}
\acmBooktitle{Proceedings of the CHI 2026 Workshop: Ethics at the Front-End: Responsible User-Facing Design for AI Systems (CHI '26), April 13--17, 2026, Barcelona, Spain}
\AtBeginDocument{%
  }

\copyrightyear{2026}
\acmYear{2026}
\acmDOI{10.48550/arXiv.2603.24853}
\setcopyright{rightsretained}
\acmConference[CHI '26]{CHI Conference on Human Factors in Computing Systems}{April 13--17, 2026}{Barcelona, Spain}
\acmISBN{}

\begin{document}

\title{Resisting Humanization: Ethical Front-End Design Choices in AI for Sensitive Contexts}

\author{Silvia Rossi}
\affiliation{%
  \institution{Immanence}
  \city{Trento}
  \country{Italy}}
\email{silvia@immanence.it}

\author{Diletta Huyskes}
\affiliation{%
  \institution{University of Milan}
  \institution{Immanence}
  \city{Milan}
  \country{Italy}}
\email{diletta.huyskes@unimi.it}

\author{Mackenzie Jorgensen}
\affiliation{%
  \institution{Northumbria University}
  \institution{Immanence}
  \city{Newcastle}
  \country{UK}}
\email{mackenzie.jorgensen@northumbria.ac.uk}

\renewcommand{\shortauthors}{Rossi et al.}

\begin{abstract}
Ethical debates in AI have primarily focused on back-end issues such as data governance, model training, and algorithmic decision-making. Less attention has been paid to the ethical significance of front-end design choices, such as the interaction and representation-based elements through which users interact with AI systems. This gap is particularly significant for Conversational User Interfaces (CUI) based on Natural Language Processing (NLP) systems, where humanizing design elements such as dialogue-based interaction, emotive language, personality modes, and anthropomorphic metaphors are increasingly prevalent. This work argues that humanization in AI front-end design is a value-driven choice that profoundly shapes users' mental models, trust calibration, and behavioral responses. Drawing on research in human-computer interaction (HCI), conversational AI, and value-sensitive design, we examine how interfaces can play a central role in misaligning user expectations, fostering misplaced trust, and subtly undermining user autonomy, especially in vulnerable contexts. To ground this analysis, we discuss two AI systems developed by Chayn, a nonprofit organization supporting survivors of gender-based violence. Chayn is extremely cautious when building AI that interacts with or impacts survivors by operationalizing their trauma-informed design principles. This Chayn case study illustrates how ethical considerations can motivate principled restraint in interface design, challenging engagement-based norms in contemporary AI products. We argue that ethical front-end AI design is a form of procedural ethics, enacted through interaction choices rather than embedded solely in system logic.
\end{abstract}



\maketitle

\section{Introduction}
Ethical discussions about AI in recent years have focused predominantly on the design and governance of back-end systems~\cite{fabric-25}, including data practices~\cite{worth2025} and mitigation of algorithmic biases~\cite{jorgensen23}. While these aspects are crucial, developers risk obscuring another ethically relevant level of AI systems: the front-end. Front-end design encompasses the interactional, linguistic, and representational choices through which users interact with AI systems, shaping not only usability but also understanding, expectations, and behavior~\cite{Ienca2023}. Despite their centrality to the user experience (UX), these design decisions are often viewed as secondary or stylistic, rather than as places where ethical values are actively negotiated. None of these design choices are neutral, but they can sometimes be shaped to protect user-related spheres~ \cite{Silva2024}.

The role of UX is particularly relevant in the context of conversational AI systems, where the proximity between people and technology is unique and adaptive. As noted by Sugisaki~\cite{Sugisaki2020}, although established principles and heuristics~\cite{Nielsen1994} “are timeless and broadly applicable to various types of user interfaces and contexts”~\cite{Sugisaki2020}, contemporary AI interfaces rely on a new paradigm characterized by “intent-based outcome specification”~\cite{Nielsen2023}. This paradigm is reflected in dialogue-based interaction, fluent natural language, emotional tone, and personalization features that make these systems appear socially responsive and similar to human interactions \cite{Kramer2025, Muehlhaus2024}. As a result, users are more likely to anthropomorphize AI systems, attributing intentions, emotions, or understandings that exceed the systems' actual capabilities. Design choices such as first-person language, describing internal processes as ``thinking'' or selectable personality modes further reinforce these perceptions~\cite{Colombatto2025, Ibrahim2025}.

To examine how ethical front-end design can respond to such anthropomorphization risks, this paper discusses two AI products for survivors from Chayn\footnote{\url{https://www.chayn.co/}}, a nonprofit organization supporting survivors of gender-based violence (GBV). In a domain characterized by heightened vulnerability, Chayn adopts a trauma-informed design approach to technology~\cite{chayn-tidp}, including their AI development~\cite{chayn-framework} that deliberately limits humanizing design cues. Rather than framing AI tools as companions or empathetic agents, the organization prioritizes transparency, clarity of scope, and user agency. This case study illustrates that ethical front-end design may, in some contexts, require resisting dominant industry trends toward engagement and human-like interaction. 

Based on this, we argue that ethical front-end design for AI should be understood as a form of procedural ethics, in which values such as autonomy, psychological integrity, and informed trust are enacted through interaction design choices rather than embedded solely in the logic of the back-end system. By foregrounding the ethical significance of user-centered design, this article contributes to ongoing discussions on human-AI interaction (HAII) ethics concerns and offers a conceptual foundation for designing human-centered AI interfaces without imitating or exploiting humans cognitively.

\section{Background}
Humanizing front-end design plays a critical role in shaping users' mental models---their beliefs about what a system is, how it works, and what can be trusted~\cite{Cena2025}. Mental models influence how users interpret system outputs, calibrate trust, and decide whether or not to trust AI-generated information. When AI systems are framed through familiar social metaphors, such as conversation or companionship, users may overestimate the system's competence, misunderstand its limitations, or assume forms of care and responsibility that the system cannot provide~\cite{Kramer2025}. Recent empirical research supports these concerns, demonstrating that AI's empathetic or warm responses can reduce system trustworthiness, increase subservient behavior, and distort users' willingness to accept or reject AI advice~\cite{Ibrahim2025}.

From an ethical perspective, these dynamics raise concerns about autonomy, transparency, and manipulation. Humanized AI interfaces can subtly exploit social norms, emotional cues, and cognitive biases, guiding user behavior in ways that are difficult to perceive or challenge~\cite{Ienca2023, Brenncke2024}. This risk is amplified in AI systems designed for personalization, microtargeting, or persuasive interaction, where front-end design becomes a key mechanism for shaping user intentions rather than simply facilitating tasks~\cite{Ienca2023}. These concerns are not evenly distributed across user populations. Individuals in vulnerable situations, such as those seeking emotional support or safety-related information, may be particularly susceptible to the effects of anthropomorphic and relational framing. For this reason, it is important to consider Value Sensitive Design approaches (VSD), “that accounts for human values in a principled and comprehensive manner throughout the design process”~\cite{Cenci2020}.

\section{Case Study: Trauma-Informed Front-End Design in AI Systems for GBV Survivors}

To illustrate how ethical considerations can actively influence AI front-end design, we examine Chayn's design of AI systems. Chayn operates in an environment characterized by their sensitive, high-stakes work. Therefore, even seemingly minor interface and interaction choices can have significant ethical implications for the internationally based survivors and allies that they support. 

From 2024-2025, Chayn collaborated with a multidisciplinary group of academic researchers and practitioners to explore the potential use of Large Language Models (LLMs) for survivors. Rather than pursuing rapid implementation or engagement-based optimization, this collaboration focused on a reflective and participatory process involving roundtable discussions, interviews, and practitioner-led conversations. As the project evolved, the focus shifted toward co-creating a framework for responsible AI development, explicitly aligned with Chayn's trauma-informed design principles~\cite{chayn-framework}. The framework includes principles-to-practice examples from Chayn's two AI systems which we discuss below. 

In 2017, Chayn deployed a chatbot called Little Window to help answer queries and point users to answers on Chayn’s website. However, in April 2020, Chayn took the product offline after reviewing chatlogs which highlighted how many distressed users misunderstood the product; users thought they were speaking to a human through Little Window, despite signposting describing this is not the case~\cite{littlewindow-takedown, ChaynBlog-littlewindow}. Since then, in 2025, Chayn deployed a LLM system called Survivor AI~\footnote{\url{https://tools.chayn.co/}} to support anyone who needs help requesting taking non-consensual images down from online platforms~\cite{ChaynBlog-survivorAI}; we note that unlike Little Window, Survivor AI is not conversational but there is still a user interface for the product which is intentionally designed. 

\subsection{Key Ethical Considerations for Front-End AI Design}

A key ethical tension in the GBV context concerns humanization. Survivors of GBV may seek empathetic and supportive interactions, but they are also disproportionately exposed to technology-mediated harm, including online harassment, impersonation, coercive control, and nonconsensual distribution of intimate content. Given this context, highly humanized or relational AI interfaces risk fostering emotional dependence, misplaced trust, or false perceptions of safety and understanding. Chayn's approach therefore deliberately resists framing AI systems as companions, friends, or quasi-human agents.

From a front-end design perspective, this resistance manifests itself in several ways. Rather than emphasizing conversational intimacy or emotional mirroring, AI-supported tools prioritize clarity of scope, explicit non-human identity, and user agency. Interaction flows are designed to support survivors' goals, such as accessing information or considering options, without encouraging personalization or emotional dependence on the system itself. This design stance contrasts sharply with dominant trends in conversational AI, where warmth, personality, and engagement are often considered default indicators of a good user experience and are associated with greater engagement. 

Importantly, Chayn's case highlights that ethical front-end design does not simply prevent harm but actively reframes success metrics. In trauma-informed contexts, reducing emotional dependence, limiting anthropomorphic cues, and maintaining epistemic humility may be ethically preferable to maximizing engagement or perceived warmth. With that said, Chayn aims to include warmth in the design of their AI systems by including calming colors to help users feel safe when interacting with their products, aligning with their safety principle~\cite{ChaynBlog-littlewindow}. Also, Chayn does not aim to maximize engagement; rather, when users engage with their AI products, they present information in non-overwhelming ways so that users move at their own pace, highlighting their agency principle. More specifically to increase user agency in the interface design, Chayn outlines the steps to users that Survivor AI takes them through and which parts involve an LLM~\cite{chayn-framework}.

\section{Conclusion}

Chayn’s adherence to their trauma-informed design principles when building AI systems demonstrates how front-end design choices can serve as a form of ethical governance, mediating the relationship between users and AI systems in ways that protect psychological integrity and autonomy. They highlight how ethical front-end design for AI can be implemented through principles of moderation, transparency, and alignment with human-centered, but not human-imitating, interaction paradigms.

\begin{acks}
The work of Silvia Rossi, Diletta Huyskes, and Mackenzie Jorgensen was funded by Immanence. 
The work of Mackenzie Jorgensen was also supported by the Engineering and Physical Sciences Re-
search Council [grant number EP/Y009800/1], through funding from Responsible Ai UK (KP0003). The authors want to thank Chayn, especially Eva Blum-Dumontet and Nadine Krishnamurthy-Spencer, and the Conversations with Practitioners working group (Kayla Evans, Sakina Hansen, Ashley Khor, Mayra Russo, and Sophia Worth) which Mackenzie Jorgensen co-led alongside Kristen M. Scott, for their impactful work bridging trauma-informed design and LLMs for the GBV space.
\end{acks}

\bibliographystyle{ACM-Reference-Format}
\bibliography{refs.bib}


\end{document}